\documentclass[10pt,twocolumn,letterpaper]{article}

\usepackage{wacv}
\usepackage{times}
\usepackage{epsfig}
\usepackage{graphicx}
\usepackage{amsmath}
\usepackage{amssymb}
\usepackage{xcolor}
\usepackage{cuted}
\usepackage{xspace}

\def\wacvPaperID{406} 

\wacvfinalcopy 

\ifwacvfinal
\def\assignedStartPage{9876} 
\fi

\ifwacvfinal
\usepackage[breaklinks=true,bookmarks=false]{hyperref}
\else
\usepackage[pagebackref=true,breaklinks=true,colorlinks,bookmarks=false]{hyperref}
\fi

\ifwacvfinal
\else
\fi

\ifwacvfinal

\newcommand{\meva}{MEVA\xspace}
\else
\newcommand{\meva}{\textit{OurDS}\xspace}

\fi

\begin{document}

\title{MEVA: A Large-Scale Multiview, Multimodal Video Dataset for Activity Detection}
\author{Kellie Corona, Katie Osterdahl, Roderic Collins, Anthony Hoogs\\
Kitware, Inc.\\
1712 Route 9, Suite 300, Clifton Park, NY 12065 USA\\
{\tt\small \{firstname.lastname\}@kitware.com}
}
\maketitle

\begin{strip}
\begin{center}
\includegraphics[width=\linewidth]{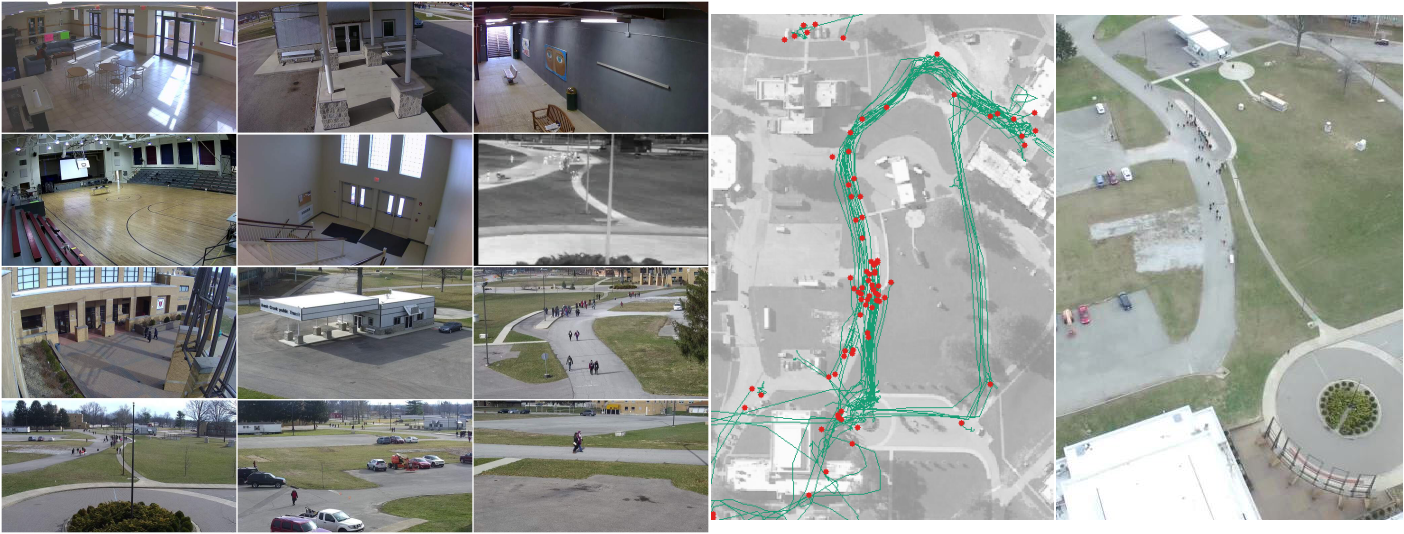} 
\fontsize{9}{10}\selectfont Figure 1. Examples of \meva data, showing an approximately synchronized view in four data modalities. Left-to-right:a 4x3 montage of RGB and thermal IR cameras; GPS locations of about 90 actors overlaid on five minutes of GPS tracks; a cropped view from a UAV.
\setcounter{figure}{1}
\label{fig:summary}
\end{center}
\end{strip}

\vspace{-0.3in}
\begin{abstract}
We present the Multiview Extended Video with Activities (\meva) dataset\cite{mevadata}, a new and very-large-scale dataset for human activity recognition. Existing security datasets either focus on activity counts by aggregating public video disseminated due to its content, which typically excludes same-scene background video, or they achieve persistence by observing public areas and thus cannot control for activity content. Our dataset is over 9300 hours of untrimmed, continuous video, scripted to include diverse, simultaneous activities, along with spontaneous background activity. We have annotated 144 hours for 37 activity types, marking bounding boxes of actors and props. Our collection observed approximately 100 actors performing scripted scenarios and spontaneous background activity over a three-week period at access-controled venue, collecting in multiple modalities with overlapping and non-overlapping indoor and outdoor viewpoints. The resulting data includes video from 38 RGB and thermal IR cameras, 42 hours of UAV footage, as well as GPS locations for the actors.  122 hours of annotation are sequestered in support of the NIST Activity in Extended Video (ActEV) challenge; the other 22 hours of annotation and the corresponding video are available on our website, along with an additional 306 hours of ground camera data, 4.6 hours of UAV data, and 9.6 hours of GPS logs. Additional derived data includes camera models geo-registering the outdoor cameras and a dense 3D point cloud model of the outdoor scene. The data was collected with IRB oversight and approval and released under a CC-BY-4.0 license. 
\end{abstract}

\section{Introduction}
\label{sec:intro}

\begin{figure*}
\begin{center}
\fontsize{9}{10}\selectfont 
\begin{tabular}{|l|l|l|l|}
\hline
person\_abandons\_package * & hand\_interacts\_with\_person & person\_reads\_document & vehicle\_picks\_up\_person\\
person\_carries\_heavy\_object & person\_interacts\_with\_laptop & person\_rides\_bicycle & vehicle\_reverses\\
person\_closes\_facility\_door & person\_loads\_vehicle & person\_puts\_down\_object & vehicle\_starts\\
person\_closes\_trunk & person\_transfers\_object & person\_sits\_down & vehicle\_stops\\
person\_closes\_vehicle\_door & person\_opens\_facility\_door & person\_stands\_up & vehicle\_turns\_left\\
person\_embraces\_person & person\_opens\_trunk & person\_talks\_on\_phone & vehicle\_turns\_right\\
person\_enters\_scene\_through\_structure & person\_opens\_vehicle\_door & person\_texts\_on\_phone & vehicle\_makes\_u\_turn\\
person\_enters\_vehicle & person\_talks\_to\_person & person\_steals\_object * & \\
person\_exits\_scene\_through\_structure & person\_picks\_up\_object & person\_unloads\_vehicle & \\
person\_exits\_vehicle & person\_purchases & vehicle\_drops\_off\_person & \\
\hline
\end{tabular}

\caption{List of the 37 activities defined and annotated in \meva. Activities marked with an asterisk are threat-based activities.}
\label{fig:act-list}
\end{center}
\end{figure*}

It has been estimated that in 2019, 180 million security cameras were shipped worldwide~\cite{cctv_market_report}, while the attention span of a human camera operator has been estimated at only 20 minutes~\cite{haering2008evolution,green1999appropriate}. The gap between the massive volume of data available and the scarce capacity of human analysts has been closed, but not eliminated, by the rapid advancement of computer vision techniques, particularly deep-learning based methods. Fundamental progress is often spurred by datasets such as ImageNet~\cite{russakovsky2015imagenet} and MS COCO~\cite{lin2014microsoft} for object recognition and MOT16~\cite{milan2016mot16} and Caltech Pedestrian~\cite{dollar2009pedestrian} for pedestrian detection and tracking.

However, as discussed in Section~\ref{sec:prior-work}, datasets for action recognition typically do not address many of the needs of the public safety community. Datasets such as AVA~\cite{Gu_2018_CVPR}, Moments in Time~\cite{monfort2019moments}, and YouTube-8m~\cite{abuelhaija2016youtube8m} present videos which are short, high-resolution, and temporally and spatially centered on the activities of interest. Rigorous research and evaluation of activity detection in public safety and security video data requires a dataset with realistic spatial and temporal scope, yet containing sufficient instances of relevant activities. In support of evaluations such as the NIST Activities in Extended Video (ActEV)~\cite{actev}, we designed the \meva dataset to explicitly capture video of large groups of people conducting scripted activities in realistic settings in multiple camera views, both indoor and outdoor. We defined 37 activity types, shown in Figure~\ref{fig:act-list}, ranging from simple, atomic single-actor instances to complex, multi-actor activities. Our fundamental resources included approximately one hundred actors on-site at an access-controlled facility with indoor and outdoor venues for a total of around three weeks, recorded by 38 ground-level cameras and two UAVs. Additionally, actors were provided with GPS loggers. We conducted extensive pre-collect planning, including a pilot collection exercise, to develop the appropriate level of actor direction required to maximize instances of our 37 activity types while maintaining realism and avoiding actor fatigue.

The final dataset contains over 9300 hours of ground-camera video, 42 hours of UAV video, and over three million GPS trackpoints. In support of the ActEV evaluation, we have annotated 144 hours of this video for the 37 activity types. While most of this data is sequestered for ActEV, we have released 328 hours of ground-camera data, 4.6 hours of UAV data, and 22 hours of annotations via the \meva website. The dataset design and collection underwent rigorous IRB oversight, with all actors signing consent forms. The data is fully releasable under a Creative Commons Attribution 4.0 (CC-BY-4.0) license, and we believe represents by far the largest dataset of its kind available to the research community.

Figure 1 shows a sample of data available from the \meva website, approximately synchronized in time, collected during a footrace scenario. The left side shows a montage of 12 of the 29 released RGB and thermal IR cameras, illustrating the diversity in locations, settings (indoors and outdoors), as well as overlapping viewpoints. The middle illustration plots approximately 90 GPS locations on a background image. The right image is a crop from UAV footage.

The paper is organized as follows. Sections~\ref{sec:prior-work} places our work in context with similar efforts. Section~\ref{sec:design} discusses how we desgined the  dataset to maximize realism and activity counts. Section~\ref{sec:annotation} describes our annotation process. Section~\ref{sec:results} briefly describes the ActEV leaderboard results on \meva.

\section{Prior Work}
\label{sec:prior-work}

We distinguish "focused" activity recognition datasets such as~\cite{Gu_2018_CVPR, monfort2019moments, abuelhaija2016youtube8m}, which typically contain single, short activities, from security-style video, which is typically long-duration and ranges from long stretches of low or no activity to busy periods with high counts of overlapping activities. Figure~\ref{fig:datasets} shows how \meva advances the state of the art along several security dataset factors, notably: duration, number of persistent fields of view (both overlapping and singular), modalities (EO, thermal IR, UAV, hand-held cameras, and GPS loggers), and annotated hours. The UCF-Crime dataset~\cite{sultani2018real} presents real incidents from real security cameras, but only from a single viewpoint at the point of activity; little or no background data is available. The VIRAT Video Dataset~\cite{CVPR2011_ViratData} contains both scripted and spontaneous activities, but without overlapping viewpoints. The Duke MTMC dataset~\cite{ristani2016performance} presents security-style video from multiple viewpoints, but with only spontaneous, unscripted activity. 

\begin{figure*}
\begin{center}
\fontsize{9}{10}\selectfont 
\begin{tabular}{|l|c|c|c|c|c|} 
\hline
\space & VIRAT~\cite{CVPR2011_ViratData} & UCF-101 Untrimmed~\cite{ucf2012wild} & Duke MTMC~\cite{ristani2016performance} & UCF-Crimes~\cite{sultani2018real} & \meva \\ [0.5ex] 
\hline\hline
Number of activity types & 23 & 101 & - & 13 & 37 \\ 
\hline
Range of samples per type & 10-1500 & 90-170 & - & 50-150 & 5-750 (1200) \\ 
\hline
Incidental objects and activities & yes & no & yes & yes & yes \\ 
\hline
Natural background behavior & yes & no & yes & yes & yes \\ 
\hline
Tight bounding boxes & yes & no & no & no & yes \\ 
\hline
Max resolution & 1920x1080 & 320x240 & 1920x1080 & 320x240 & 1920x1080 \\ 
\hline
Sensor modalities & 1 & 1 & 1 & 2 & 5 \\ 
\hline
Security & yes & no & yes & yes & yes \\ 
\hline
Number of FOV & 17 & unique-per-clip & 8 & unique-per-clip & 28 \\ 
\hline
Overlapping FOV & no & no & yes & no & yes \\ 
\hline
Indoor \& outdoor & no & yes & no & yes & yes \\ 
\hline
Availability & direct & reference & \textit{retracted} & direct & direct \\ 
\hline
Clip length & 2-3 minutes & 1-71 seconds & - & 4 minutes & 5 minutes \\ 
\hline
Dataset duration (hours) & 29 & 26.6 & 10 & 128 & 9300 / 144 \\
\hline
\end{tabular}

\caption{ Comparison of characteristics of activity and security datasets to \meva. The reported range of activity counts per type is only for \meva released data while the average number of samples in parenthesis includes both released and sequestered annotations. The 5 sensor modalities included in the \meva dataset are static EO, UAV EO, Thermal IR, Handheld or Body-Worn Cameras (BWC), and GPS. There are 28 unique FOV with additional FOVs offered by drone, handheld and BWC footage. The overall duration of video collected as part of \meva is 9300 hours while 144 hours were annotated.}
\label{fig:datasets}
\end{center}
\end{figure*}

\section{Dataset Design}
\label{sec:design}

\begin{figure}
\begin{center}
\includegraphics[width=\linewidth]{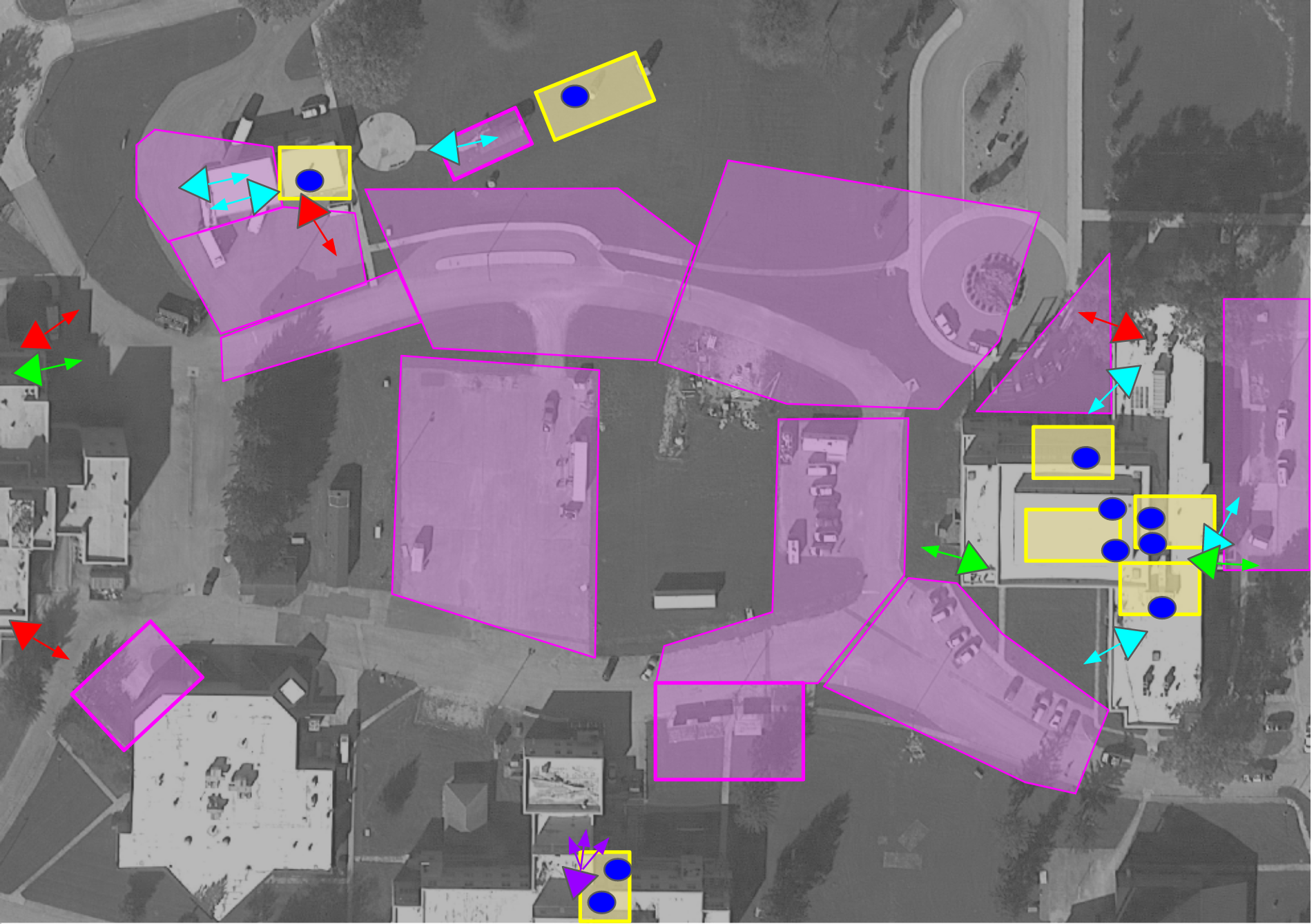}
\caption{Site map with released cameras and approximate fields-of-view. Indoor cameras are blue circles; triangles are outdoor cameras. Red are co-located EO/IR; blue is fixed-focal-length; green is PTZ in stationary mode; purple is PTZ in patrol mode. Pink fields of view are outdoors; yellow are indoors. }
\label{fig:site-map}
\end{center}
\end{figure}

The \meva dataset was designed to capture human activities, both spontaneous and scripted, in an environment as close as possible to that encountered by deployed CCTV systems in the real world. The dataset was designed to be realistic, natural and challenging for security and public safety video research in terms of its resolution, background clutter, visual diversity, and activity content. As discussed in Section~\ref{sec:release}, the data is fully releasable under a Create Commons Attribution 4.0 (CC-BY-4.0) license, facilitated by a detailed human subjects research plan. Activity diversity was achieved through pre-collect scripting for scenarios and activity types (Section-\ref{sec:script}), and collected via an ambitious camera plan with 38 ground-level cameras and two UAVs (Section-\ref{sec:collect}). Scene diversity was accomplished by scripting scenarios to occur at different times of day and year to create variations in environmental conditions, and directing demographically diverse actors to perform activities in multiple locations with varied natural behaviors. The ground-camera plan includes a variety of camera types, indoor and outdoor viewpoints, and overlapping and singular fields of view. Realism was enhanced by scripting to include naturally occurring activity instances, frequent incidental moving objects and background activities, as discussed in Section-\ref{sec:script}. Finally, the quantity of data collected was enabled by planning for a multi-week data collection effort to allow for a large set of activity types with a large number of instances per activity class as discussed in Section-\ref{sec:collect}.

\subsection{Releasability}
\label{sec:release}
A critical requirement for the \meva dataset is broad releasablility to the research community. Possible restrictions include prevention of commercial use, which discourages commercial companies from using the dataset for research or collaborating on teams which do use the dataset. Another restriction is when data containing human activity is collected without oversight of an Institutional Review Board (IRB) or collected outside of its initially IRB-approved conditions. While the data may address other needs of the research community, lack of proper IRB oversight or participant consent can render the data unusable, wasting significant resources in the collection and curation process. 

The \meva dataset was carefully crafted to avoid such restrictions to maximize its value to the research community. The data collection effort was overseen by an IRB to ensure that the data collected would be usable. We wrote a human subjects research protocol and informed consent documents that detailed the collection process, what types of data would be collected, how the data will be released and may be used, and the risks and benefits to anyone consenting to participate in the study. The informed consent document was provided to prospective participants in advance and reviewed in an informed consent briefing session prior to beginning the data collection. Included in the protocol and consent briefing was a sample license that may be used for the data, a CC-BY 4.0, to make it fully clear the intent and potential future use of the data. The consent presentation was followed by small group discussions and question/answer periods to thoroughly address and answer any participant concerns prior to them consenting to participate the research. Once the individuals who voluntarily chose to participate signed the consent forms, we were able to begin the data collection process. 

To further ensure broad releasability of the data, the collection occurred at an access-controlled facility to minimize non-participating individuals entering the fields of view of the cameras and appearing in the data. 

\subsection{Scripting}
\label{sec:script}
To guarantee collection of minimum activity instance counts, ensure diversity in the data, and produce the most realistic video data, \meva data was collected for both scripted and unscripted activities. The scripting challenges for the data collection effort can broadly be broken down into two categories: (1) scripting for satisfying program requirements such as quantity and diversity, and (2) scripting for realism in activity behavior.

It was critical to design the data collection to represent a variety of realistic scenarios in ground-based security video collection. We determined seven overarching scenarios of interest to the public safety community. For example a footrace with multiple phases including registration, participant arrival, race and cleanup. The scenarios were collected over a period of three weeks on different days and at different times of days to capture video variations due to weather and lighting. At the extreme, two sub-collects in March and May produced dramatic contrast in weather from snow to extreme heat, and thus diversity in natural human behavior (e.g., clustering together in the cold) and wardrobe (i.e., jackets in the winter and t-shirts in the summer).

\begin{figure}[t]
\begin{center}
\includegraphics[width=\linewidth]{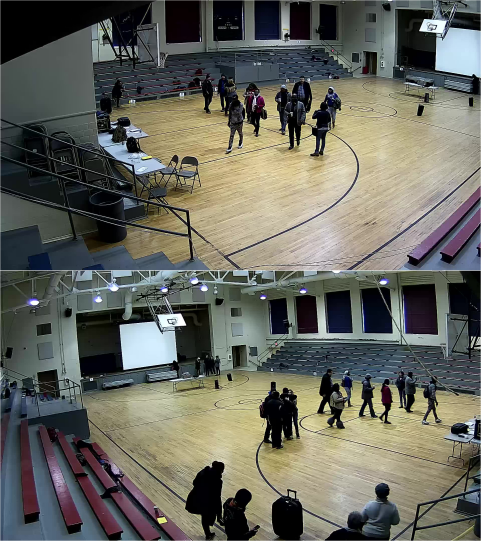}

\caption{Two approximately synchronized views from the gym.}
\label{fig:indoor1}
\end{center}
\end{figure}

Scripted into these scenarios were 37 primitive, complex and threat-based human activity types to ensure minimum instance counts useful to the program; these are listed in Figure~\ref{fig:act-list}. These activities were scripted to be performed by actors with different age, sex and ethnicity as allowed by the actor pool. Activities involving vehicles or props were specifically scripted to rotate the vehicle pool and theatrical property. Scripted activities occurred across various scenarios and at multiple locations in the facility. Activities were scripted in overlapping and singular fields of view, indoor and outdoor cameras, and various camera types (e.g., stationary EO and IR, roaming EO). Activities were scripted to be performed by the same individual in different, singular cameras with the goal of associating the same individual across different cameras. This is relevant when a central activity is detected involving a subject, and then a derived complex activity is to detect other individuals that meet with the original subject of interest. Scenarios also included confusers for the scripted activity types to add value for performer evaluation in the final dataset. Figure~\ref{fig:indoor1} shows two approximately synchronized views from the gym.

Though scenarios had an overarching theme, multiple iterations of the same scenario (called takes) were designed to produce visibly distinct but comparable results by varying the parameters described above. Additionally, actor density at various location in the facility and traffic flow were modified to produce differences in subsequent takes of the same scenario. Scripting the scenarios and activities in this level of detail guaranteed the diversity and quantity of data satisfied program requirements.

To address the second broad goal, we hired professional actors to act scenarios in a natural manner to create an equivalent challenge for computer vision event detectors, as if the events occurred in real-life scenarios without acting. The actors were divided into squads which each had a squad lead in charge of managing a group of 5-8 individuals. The squads were designed, grouped and shuffled to produce variations in demographics, behaviors and inter-actor behaviors. Each squad was given a direction card which provided information on timing and locations for activities to be performed, and group behavior such as social groupings within the squad. By giving the squad lead the ability to designate roles within the group, natural social dynamics, such as those observed in married couples or families, were preserved and increased the realism of the collected data. The squad lead was responsible for assigning roles to the actors, managing vehicle and prop assignment, providing feedback to individual actors for corrections on activity behavior, and reporting descriptions of complex or threat-based activities for use in expediting later annotation. 

We found that providing scripting at the squad rather than individual level produced the most realistic actor behavior. When actor behavior was scripted on the individual level at high temporal and spatial resolution, the actors, overwhelmed by the details, focused on achieving all activities at the specified time and location rather than performing activities that were natural in appearance. Using a small pilot collect of data, we were able to determine which activities occurred naturally in the data collection without direction (e.g., person\_opens\_door). Taking this into account when scripting, lower instances of these activities were scripted and a mix of scripted and unscripted occurrences were collected. On the other hand, rare and threat-based activities needed to be specifically scripted to ensure minimum counts of these activities were achieved and later located for annotation. However, allowing the actors to have artistic licence with how to accomplish activities, especially complex activities, produced more a natural and diverse dataset. Actor  familiarity with the scripted scenario also increased realism. Most scenarios varied in time from 15 minutes to 1 hour. Multiple takes could be run consecutively with minor modifications to scripted behaviors and activity instances, and minimal reset time (5 to 10 minutes) to efficiently produced varied instances of a scenario and communicate modifications to the squad leads.

In addition to activity-focused scripting, scene-level behaviors also needed to be scripted to ensure realism. For example, train and bus arrival schedules were designed to reflect public transit schedules scaled for the population depicted by our actors. Vehicle traffic was scripted to have ebbs and flows associated with similar patterns to real-world versions scripted scenarios. For example, an increase in vehicles dropping off people at the bus station just prior to a bus's scheduled arrival. Driving routes, vehicles and drivers were shuffled to provide variations between different scenarios and takes. Again, these tasks were assigned at the squad level and managed by squad leads.

In addition to scripted scenarios, periods of completely unscripted data were collected with the aim of collecting naturally occurring common activities with no actor direction. We also took advantage of unscripted data to interject threat-based and rare activities to capture undirected response by casual observers. These unscripted times were naturally occurring and added value to reset times between scenario takes.

\subsection{Collection}
\label{sec:collect}

\begin{figure}
\begin{center}
\includegraphics[width=\linewidth]{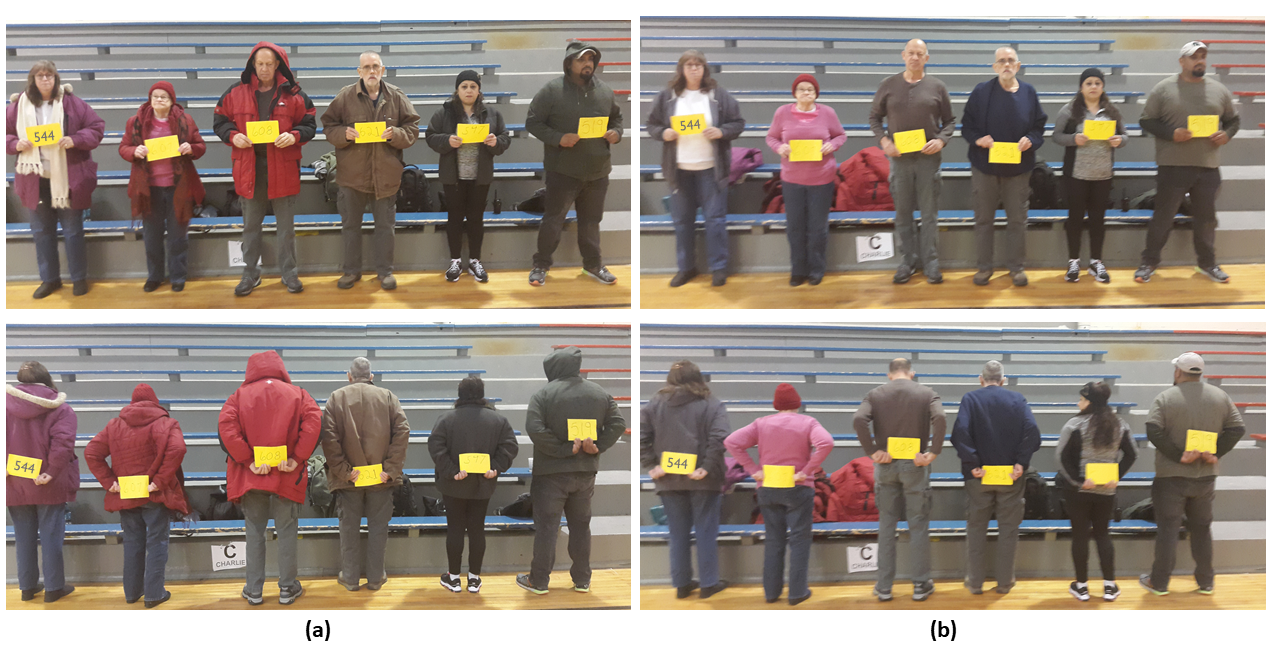}
\caption{Enrollment photos were taken daily of all actors, with (a) and without (b) outerwear, from the front and back. Photos included actor numbers, which correspond to their unique GPS logger number for reidentification purposes. }

\label{fig:enrollmentfig}
\end{center}
\end{figure}

The \meva video dataset was collected over a total of three weeks at the Muscatatuck Urban Training Center (MUTC) with a team of over 100 actors and 10 researchers. MUTC is an access-controlled facility run by the Indiana National Guard that offers a globally unique, urban operating environment. The real and operational physical infrastructure, including curbed roads and used buildings, set it apart from other access-controlled facilities for collecting realistic video data.

The camera infrastructure included commercial-off-the-shelf EO cameras; thermal IR camerasas part of several EO-IR pairs, two DJI Inspire 1 v2 drones, and a range of video and still images from handheld cameras used by the actors. The fields of view, both overlapping and non-overlapping, capture person and vehicle activities in indoor and outdoor settings. 

Onsite staging and actor instruction also improved the realism in scenes for the \meva scenario collection. Areas were staged with props, such as signs, tables, chairs, and trashcans, to provide scenario-specific context necessary to increase the natural appearance and usefulness of the area to both actors and recording cameras. Additionally, areas were staged to look similar but distinct between different scenarios to add diversity to the dataset collected. For example, the foyer of one building was converted into an operational cafe where the actors were able to grab drinks and snacks during scripted or unscripted camera time.

\begin{figure}
\begin{center}
\includegraphics[width=\linewidth]{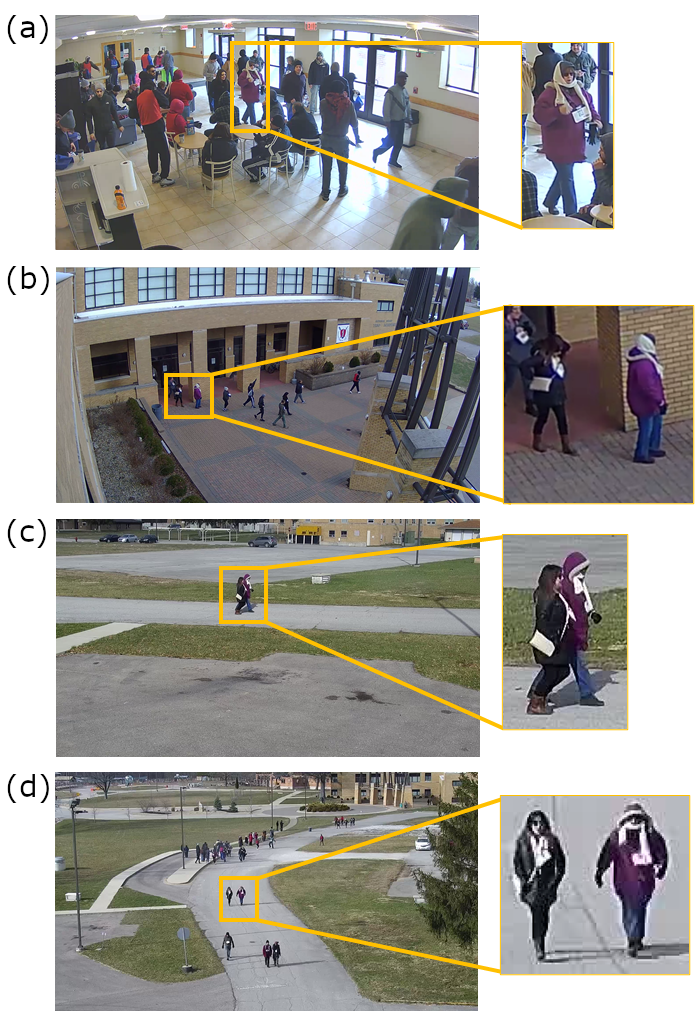}
\caption{The actor in the purple jacket, actor 544 in Figure~\ref{fig:enrollmentfig}, is visible in multiple cameras during the scenario. Her height is (a) 301 pixels, (b) 118 pixels, (c ) 176 pixels, and (d) 89 pixels in each of the respective fields of view.}

\label{fig:reidfig}
\end{center}
\end{figure}

Actors were selected to provide a diverse pool of individuals in background, ethnicity and gender. As part of our iterative process to receive and incorporate feedback, all actors attended briefings in which camera fields of view and common issues (e.g., erratic driving or aimless walking) were discussed in a group setting, raising awareness of key factors required to collaboratively produce realistic behaviors in video. Squad leads also held smaller briefings, providing feedback to the scripting team and receiving instructions to pass to their squad members. During morning briefings, all actors and \meva team members were registered with a unique identifying number matched to their GPS logger unit and wardrobe enrollment photograph(s) as shown in Figure~\ref{fig:enrollmentfig}. The photos, paired with the GPS logs, enable reidentification of individual actors across fields of view in a single scenario, as seen in Figure~\ref{fig:reidfig}. Each day before filming, squad leads were given direction cards to assign roles to their actors and inform prop distribution, then data collection was performed. Shorter scenario takes of less than 1 hour were performed in rapid succession with only minor modifications relayed during a brief reset. One day short-scenario data collection would contain between 4 and 8 takes; for longer scenarios of approximately 8 hours, the scenario would be repeated across multiple sequential days. Direction and course correction was provided by both directly interjecting a \meva team member into the scenario for a brief interval or incorporating actor briefings naturally into the scenario schedule.

The final dataset contains 9,303 hours of ground-camera video (both EO and thermal IR), 42 hours of UAV footage, and 46 hours from hand-held and body-worn cameras. Additionally we collected over 2.7 million GPS trackpoints from 108 unique loggers.
\section{Annotation}
\label{sec:annotation}

Annotating a large video dataset requires balancing quality and cost.   Annotating for testing and evaluation, as is the case for \meva, must prioritize the quality of ground truth annotations, as any evaluation will only be as reliable as the ground truth derived from the annotations. Ground truth annotations must minimize missed and incorrect activities to reduce the potential for penalizing correct detections, strictly adhere to activity definitions, and address corner cases to reduce ambiguity in scoring.  We annotated \meva by first localizing spatio-temporal activities and then creating bounding boxes for objects involved in the activity. We optimized a multi-step process for quality and cost: (1) an annotation step to identify the temporal bounds for activities and objects involved, (2) an audit step by experts to ensure completeness and accuracy of the activity annotations, (3) a crowdsourced method for bounding box track creation for objects, and (4) a custom interface for rapid remediation of corner cases by experts, with automated checks for common human errors between each of these steps.

\subsection{Activity Annotation}
\label{sec:activity}

\begin{figure}
\begin{center}
\includegraphics[width=\linewidth]{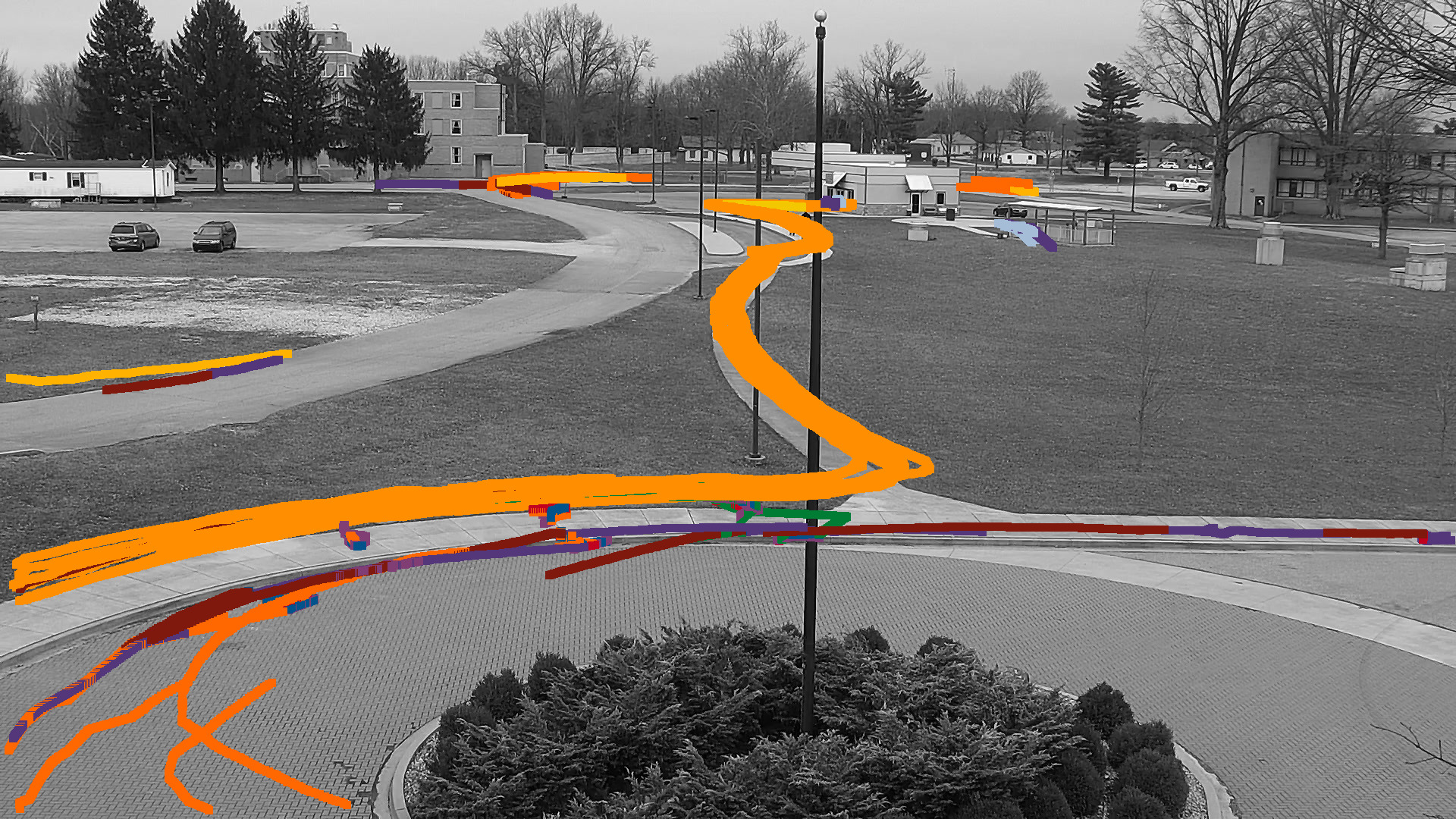}
\caption{Half an hour of activities overlaid on a background image. 112 activities, 17 types, ranging from 1 to 14 instances.}
\label{fig:outdoor-half-hour}
\end{center}
\end{figure}

We annotated \meva through a sequential process of activity identification by a trained, dedicated team of third-party annotators, followed by a quality control audit from an internal team of \meva experts intimately involved in defining the activities.  The initial activity identification step detects all \meva activities in a 5 minute video clip, specifying the start and end times for each activity, and identifying the initial and final location for each actor or object involved.  The subsequent audit step catches any missed or incorrectly labeled activity instances, and confirms strict adherence to \meva definitions. 

Ambiguity in activity definitions is a fundamental annotation challenge. To minimize this, we define activities very explicitly and require that events are always annotated on pixel evidence and not human interpretation.  We developed detailed annotation guidelines for each activity, including definition of the start and end times for an activity, and the actors and objects involved. Definitions also include extensive discussion of corner cases. 

We explored several methods to increase annotation efficiency and quality, including completely crowdsourced methods on Amazon Mechanical Turk (AMT) and a solely in-house team of experts. The optimized annotation process used a team of third-party annotators dedicated to \meva annotation to lower the overall cost and supply surge capacity via dynamic team scaling while still guaranteeing quality results. After comparing  the results of multiple annotators performing activity annotation in parallel and combined results in a post-processing step with annotating in series with a quality assurance step, serial annotation was selected to enable efficiently annotating a larger dataset. 

Annotation consistency is essential to providing high-quality ground truth annotations.  When using a diverse team of annotators, it was essential to have a clear set of guidelines and iterative definition modifications to include corner cases to ensure annotator agreement across the team. Additionally, we found that project-specific training, including effective use of the annotation tools, in-depth discussion of definitions including examples, and iterative feedback on annotation quality produced improvements in annotation quality and speed. Finally, a quality audit from an internal team of \meva annotators was used to guarantee no missed or incorrect instances.  We observed a 3-fold increase in audit efficiency over a two month period of working in a tight feedback loop with these procedures and a project-dedicated team of annotators. 


\subsection{Track Annotation}
\label{sec:track}
Once activities and participating objects were identified by \meva experts, objects were tracked with bounding boxes for the duration of their involvement in the activity. The annotation of bounding boxes was primarily conducted through crowdsourced annotation on AMT, via an iterative process of bounding box creation and quality review.

In the first step of bounding box refinement, videos were broken up into segments that corresponded to a single activity annotation. An AMT task was created for each activity, displaying a set of start and end boxes with linearly interpolated boxes on intervening frames for all objects involved in the activity. Workers were instructed to complete the tracks by annotating bounding boxes for the interpolated frames which were as tight as possible around the object; for example, all the visible limbs of a person should be in the bounding box while minimizing "buffer" pixels. These two characteristics are required in the \meva dataset to ensure that high-quality activity examples with minimal irrelevant pixels are provided for testing and evaluation.  These traits were enforced by using other AMT annotators to assess the quality of the resulting tracks.  

In the quality review step, the bounding box annotations were shown to several AMT workers who were asked to evaluate them as acceptable according our guidelines. If agreement was achieved between the AMT workers, then the results were considered acceptable and complete; otherwise, a new AMT task would be created for bounding box refinement. The results of the next round of refinement would be provided for quality review, and the process would repeat until acceptable annotations were produced or a threshold number of iterations were performed. If the threshold number of iterations were performed, the clips would be vetted and edited manually as needed by an \meva auditor.  The percentage of activities requiring expert intervention was less than 5\%. Specialized web interfaces were developed for each of these tasks to allow workers (AMT or in-house) to easily provide the necessary results. In order to eliminate most systematic low quality jobs, annotations were sampled and workers were allowed to continue based on quality.

\begin{figure}
\begin{center}
\includegraphics[width=\linewidth]{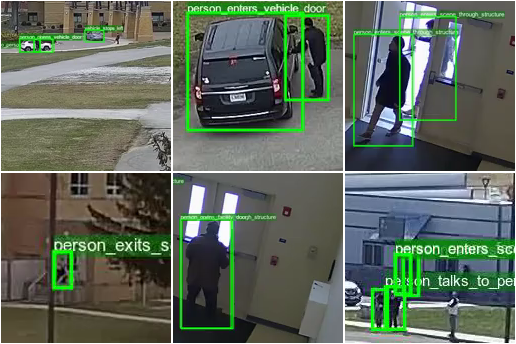}
\caption{Examples of activities and tracks from diverse fields of view. The font size in each images is consistent, indicating varying scale of the annotations.}
\label{fig:tracks}
\end{center}
\end{figure}

Figure~\ref{fig:tracks} illustrates bounding box annotations for tracks from a variety of activities and fields of view that demonstrate the quality produced by our track annotation procedure. Box size varied dramatically due to the scale variation in the video, with a mean area of 13559$\pm$23799. Annotations span 5 track types (person, vehicle, other, bag, and bicycle) with a distribution of 90.71\%, 4.5\%, 4.51\%, 0.15\%, and 0.05\%, respectively. As part of quality control, we compared \meva annotations against high-confidence performer false alarms from the ActEV evaluation; our false negative rate (i.e., confirmed missed instances to total activity count in reviewed clips) is less than 0.6\% . 

\begin{figure}
\begin{center}
\includegraphics[width=\linewidth]{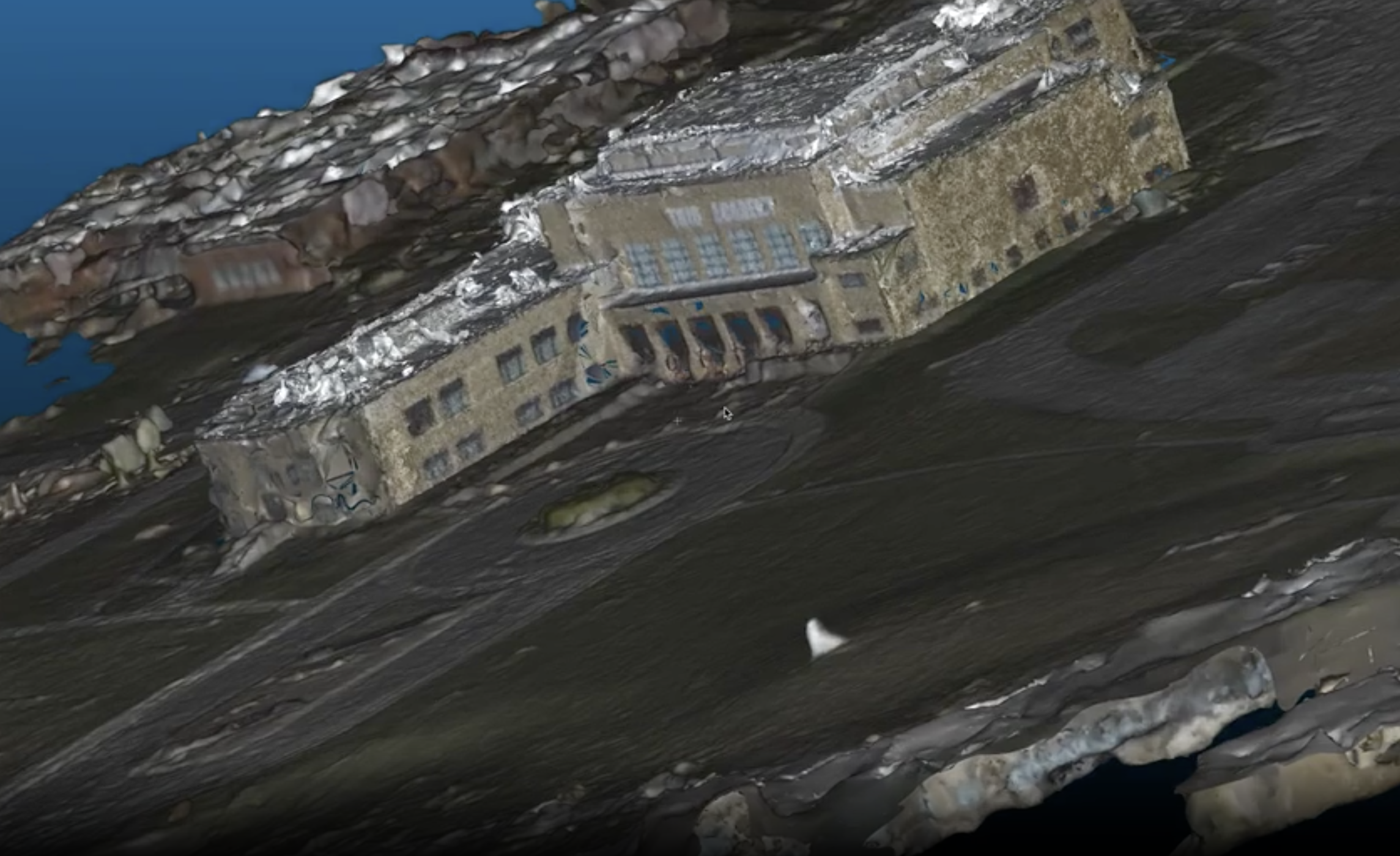}
\caption{Visualization of the fine-grained 3D point cloud model of the collection site.}
\label{fig:3d-model}
\end{center}
\end{figure}

\textbf{Additional Data} In addition to the annotations, we have provided supplemental data such as camera models for the camera models which register into a common geo-referenced coordinate system. We have also provided a 3D model of the outdoor component of the collection site, provided as a PLY file and visualized in Figure~\ref{fig:3d-model}.

\section{Baseline Activity Detection Results}
\label{sec:results}

\begin{figure}
\begin{center}
\includegraphics[width=\linewidth]{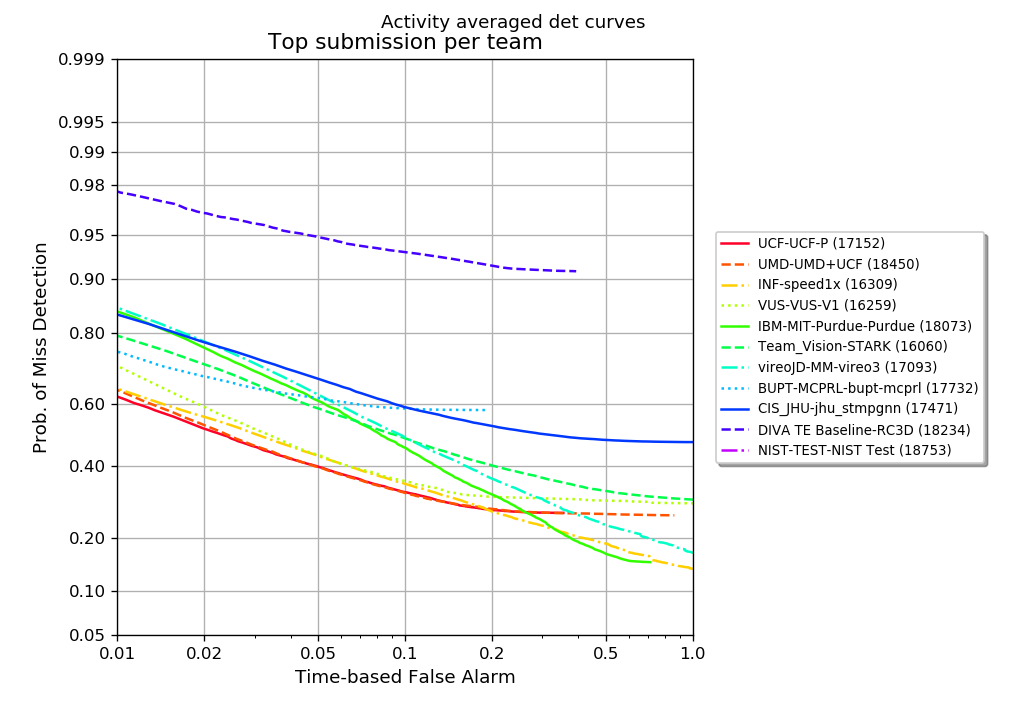}

   \caption{The NIST ActEV leaderboard as of 9 Nov 2020, computed on \meva data. Better performance is lower and to the left.}
\label{fig:actev-graph}
\end{center}
\end{figure}

The NIST ActEV challenge~\cite{actev} is using the \meva dataset for its ongoing Sequestered Data Leaderboard (SDL). The ActEV challenge defines three tasks: Activity Detection (AD) with no spatial localization in the video, Activity and Object Detection (AOD) where the activity and participating objects are spatially localized within a frame but not necessarily correlated across frames, and Activity / Object Detection and Tracking (AODT), which extends AOD to establish real-world activity and object identity across frames. The current leaderboard is for AD, scored using Probability of Miss (pMiss), the proportion of activities which were not detected for at least one second, vs. Time-based False Alarm (TFA), the proportion of time the system detected an activity when there was none. Figure~\ref{fig:actev-graph} shows results current at time of writing for nine teams plus a baseline implementation~\cite{actev-baseline} based on RC3D~\cite{xu2017r}. As the scores indicate, the \meva dataset  is very difficult compared to related datasets, presenting abundant opportunities for innovation and advancement in activity detection, tracking, re-identification and other problems.

\section{Conclusion}
\label{sec:conclusion}

We have presented the \meva dataset, a large-scale, realistic video dataset  containing annotation of a diverse set of visual activity types.  The \meva video dataset surpasses existing activity detection datasets in hours of video, number of cameras providing overlapping and singular fields of view, variety of sensor modalities, and broad releasability. The dataset also provides a substantial 144 hours of evaluation-quality activity annotations of scripted and naturally occurring activities. We believe that with these traits the \meva dataset will stimulate diverse research within the computer vision community. The \meva dataset is available at: \url{http://mevadata.org}.

\textbf{Acknowledgement:} This work is supported by the Intelligence Advanced Research Projects Activity (IARPA) via contract 2017-16110300001. The U.S. Government is authorized to reproduce and distribute reprints for Governmental purposes notwithstanding any copyright annotation thereon. Disclaimer: The views and conclusions contained herein are those of the authors and should not be interpreted as necessarily representing the official policies or endorsements, either expressed or implied, of IARPA or the U.S. Government.

{\small
\bibliographystyle{ieee_fullname}
\bibliography{dataset_bib}
}

\end{document}